\documentclass[11pt]{article}

\usepackage[final]{acl}

\usepackage{amssymb}
\usepackage{array}
\usepackage{xspace}
\usepackage{tabularx}
\usepackage{adjustbox}
\usepackage{booktabs}
\usepackage{multicol}
\usepackage{multirow}
\usepackage{graphicx}
\usepackage{listings}
\usepackage[table]{xcolor}
\usepackage{makecell}
\usepackage{pifont}

\usepackage{times}
\usepackage{latexsym}

\newcommand{\cmark}{\textcolor{green!70!black}{\ding{51}}}
\newcommand{\xmark}{\textcolor{red!80!black}{\ding{55}}}

\title{A Survey on MLLM-based Visually Rich Document Understanding: Methods, Challenges, and Emerging Trends}

\author{
 \textbf{Yihao Ding\textsuperscript{1}},
 \textbf{Siwen Luo\textsuperscript{1}} \thanks{Corresponding Author.},
 \textbf{Yue Dai\textsuperscript{2}},
 \textbf{Yanbei Jiang\textsuperscript{2}},
 \textbf{Zechuan Li\textsuperscript{2}}, \\
 \textbf{Qiang Sun\textsuperscript{1}},
 \textbf{Geoffrey Martin\textsuperscript{3}},
 \textbf{Wei Liu\textsuperscript{1}},
 \textbf{Yifan Peng\textsuperscript{3}}
\\
 \textsuperscript{1}The University of Western Australia, Crawley, Australia\\
 \textsuperscript{2}The University of Melbourne, Melbourne, Australia\\
 \textsuperscript{3}Weill Cornell Medicine, New York, USA
\\
 \small\textbf{Correspondence:} \texttt{\{yihaoding, siwen.luo\}@uwa.edu.au}
}
\begin{document}
\maketitle

\begin{abstract}
Visually Rich Document Understanding (VRDU) has become a pivotal area of research, driven by the need to automatically interpret documents that contain intricate visual, textual, and structural elements. Recently, Multimodal Large Language Models (MLLMs) have demonstrated significant promise in this domain, including both OCR-based and OCR-free approaches for information extraction from document images. This survey reviews recent advances in MLLM-based VRDU, highlighting emerging trends and promising research directions with a focus on two key aspects: (1) techniques for representing and integrating textual, visual, and layout features; (2) training paradigms, including pretraining, instruction tuning, and training strategies. Moreover, we address challenges such as data scarcity, handling multi-page and multilingual documents, and integrating emerging trends such as Retrieval-Augmented Generation and agentic frameworks. Our analysis offers a roadmap for advancing MLLM-based VRDU toward more scalable, reliable, and adaptable systems.

\end{abstract}

\section{Introduction}

Visually-Rich Document Understanding (VRDU) lies at the intersection of vision and language, aiming to extract and understand information from documents with multiple data modalities and complex layouts \cite{cord, formnlu}. With the rapid digitization of physical documents and the widespread use of structured and semi-structured digital documents, the development of robust, generalizable VRDU frameworks has attracted significant attention for automating information extraction, improving accessibility, and enhancing decision-making across diverse domains such as finance, healthcare, and education.
{\begin{table*}[t]
\scriptsize
\centering
%\rowcolors{2}{gray!10}{white}
\begin{adjustbox}{width=\textwidth}
\setlength{\tabcolsep}{4pt}
\begin{tabular}{llllllccccl}
\toprule
\textbf{Model} & \textbf{Venue} & \textbf{Tasks} & \textbf{Mod.} & \textbf{LLM Backbone} & \textbf{Vision Encoder} & \textbf{PT} & \textbf{IT} & \textbf{FT} & \textbf{Pages} & \textbf{Prompt In.} \\
\midrule
\rowcolor[gray]{.9}\textbf{OCR-Dependent}&&&&&&&&&&\\
ICL-D3IE \shortcite{icld3ie}        & ICCV    & KIE        & T, L      & GPT-3          & --             & \xmark & \xmark & \xmark & SP & ICL+Layout   \\
DocLLM \shortcite{docllm}           & ACL     & KIE, QA, DC& T, L      & Custom         & --             & \cmark & \cmark & \xmark & SP & T+B+Q        \\
LAPDoc \shortcite{lapdoc}           & ICDAR   & KIE, QA    & T, L      & Multiple       & --             & \xmark & \xmark & \xmark & SP & Rule         \\
LMDX \shortcite{lmdx}               & ACL     & KIE        & T, L      & Gemini-pro     & --             & \xmark & \xmark & \xmark & SP & ICL+Layout   \\
ProcTag \shortcite{proctag}         & AAAI    & QA         & T, V, L      & GPT-3.5        & --             & \xmark & \xmark & \cmark & SP & Rule+CoT     \\
DocKD \shortcite{dockd}             & EMNLP   & KIE, QA, DC& T, L      & Custom            & --             & \xmark & \xmark & \cmark & SP & Gen by VL    \\
DoCo \shortcite{li2024enhancing}    & CVPR    & KIE, QA, DC& T, L      & Multiple       & LayoutLMv3      & \cmark & \xmark & \cmark & SP & I+Q          \\
InstructDoc \shortcite{instructdoc}  & AAAI    & KIE, QA    & T, V, L   & FlanT5         & LayoutLMv3      & \xmark & \cmark & \cmark & MP & I+Q          \\
LayoutLLM \shortcite{luo2024layoutllm} & CVPR  & KIE, QA    & T, V, L   & Vicuna-7B-v1.5 & LayoutLMv3  & \xmark & \cmark & \cmark & SP & I+Q+CoT      \\
LLaVA-Read \shortcite{llavaread}    & preprint& KIE, QA    & T, V, L   & Vicuna-1.5 13B & Multiple             & \cmark & \cmark & \xmark & SP & I+Q          \\
LayTextLLM \shortcite{lu2024bounding} & ACL& QA, KIE  & T, L      & Llama2-7B-base & --             & \cmark & \xmark & \cmark & SP & T+B          \\
DocLayLLM \shortcite{doclayllm}     & CVPR    & QA, KIE    & T, V, L   & Llama2-7B-chat & Pix2Struct-Large& \xmark & \cmark & \cmark & SP & I+Q+B        \\
LayTokenLLM \shortcite{laytokenllm} & CVPR    & QA         & T, L      & Multiple       & --             & \cmark & \xmark & \xmark & MP & I+Q+L        \\
GPE \shortcite{gpe}                 & ICLR    & KIE, QA    & T, L      & Multiple          & --             & \xmark & \xmark & \cmark & SP & T+B+Q        \\
MDocAgent \shortcite{mdocagent}     & preprint& QA         & T, V      & Multiple       & ColPali, ColQwen2    & \xmark & \xmark & \xmark & MP & I+Q          \\
PDF-WuKong \shortcite{pdfwukong}    & preprint& QA         & T, V      & BGE-M3         & IXC2-VL-4KHD      & \xmark & \xmark & \cmark & MP & I+Q          \\

DocAssistant \shortcite{zhang2025docassistant} & EMNLP & QA & T, V & InternVL2-Chat-2B & InternVL2 ViT &  \xmark & \xmark & \cmark & SP & I+Q\\

%SimpleDoc \shortcite{jain2025simpledoc} & EMNLP & QA & T, V & \\

%MultiDocFusion \shortcite{shin2025multidocfusion} & EMNLP\\

AlignVLM \shortcite{masryalignvlm} & Neurips & QA & T, V & LLaMA-3.2 (1B, 3B) & SigLIP-400M & \cmark & \cmark & \cmark &  SP & I+Q\\

DocThinker \shortcite{yu2025docthinker} & ICCV & QA, KIE & T, V & Qwen2.5-VL (3B, 7B) &Qwen2.5-VL ViT & \xmark & \xmark & \cmark & SP & I+Q\\
\midrule

\rowcolor[gray]{.9}\textbf{OCR-Free}&&&&&&&&&&\\
KOSMOS-2.5 \shortcite{lv2023kosmos}    & preprint & QA, KIE     & V          & Custom         & mPLUG-Owl VE     & \xmark & \cmark & \cmark & SP & I+Q          \\
mPLUG-DocOwl \shortcite{mplugdocowl}   & preprint & QA          & V          & mPLUG-Owl      & mPLUG-Owl VE     & \xmark & \cmark & \xmark & SP & I+Q          \\
UReader \shortcite{ureader}            & EMNLP    & QA          & V          & mPLUG-Owl      & mPLUG-Owl VE     & \xmark & \cmark & \xmark & SP & I+Q          \\
TGDoc \shortcite{wang2023towards}      & preprint & KIE, QA     & V          & Vicuna-7B      & CLIP-ViT-L/14    & \xmark & \cmark & \cmark & SP & I+Q+B        \\
UniDoc \shortcite{unidoc}              & preprint & KIE, QA     & V          & Vicuna-7B      & CLIP-ViT-L/14    & \xmark & \cmark & \cmark & SP & I+Q+B        \\
DocPedia \shortcite{feng2024docpedia}  & SCIS     & KIE, QA     & V          & Vicuna-7B      & Swin Trans.      & \cmark & \xmark & \cmark & SP & I+Q          \\
HRVDA \shortcite{hrvda}                & CVPR     & KIE, QA     & V          & LLama2-7B      & Swin Trans.      & \cmark & \cmark & \xmark & SP & I+Q          \\
Vary \shortcite{vary}                  & ECCV     & QA, DocRead & V          & Multiple          & CLIP, ViTDet     & \cmark & \xmark & \cmark & SP & I+Q          \\
mPLUG-DocOwl1.5 \shortcite{mplugdocowl15}& EMNLP  & KIE, QA     & V          & mPLUG-Owl2     & mPLUG-Owl2 VE    & \xmark & \cmark & \cmark & SP & I+Q          \\
HVFA \shortcite{park2024hierarchical}  & Neurips     & QA, Cap.    & V          & Multi (BLIP-2, etc.)& ViT/L-14     & \xmark & \cmark & \xmark & SP & I+Q          \\
Texthawk \shortcite{texthawk}          & preprint & QA          & V          & InternLM-XC    & ViT              & \xmark & \cmark & \cmark & SP & I+Q          \\
Texthawk2 \shortcite{texthawk2}        & preprint & OCR, Grd, QA& V          & Qwen2-7B-Instr & SigLIP-SO400M    & \xmark & \cmark & \cmark & MP & I+Q+Task     \\
TextMonkey \shortcite{textmonkey}      & preprint & KIE, QA     & V          & Qwen-VL        & Vit-BigG         & \xmark & \cmark & \xmark & SP & I+Q          \\
Llavar \shortcite{llavar}              & preprint & QA          & V          & Vicuna-13B     & CLIP-ViT-L/14    & \xmark & \cmark & \cmark & SP & I+Q          \\
TokenCorrCompressor \shortcite{zhangtoken}& preprint & QA, Cap. & V          & LLaMA-2        & CLIP-ViT/L14     & \xmark & \xmark & \cmark & SP & I+Q          \\
DocKylin \shortcite{dockylin}          & AAAI     & QA          & V          & Llama2-7B-chat & Donut-Swin       & \xmark & \cmark & \cmark & SP & I+Q          \\
Marten \shortcite{marten}              & CVPR     & QA          & V          & InterLM2       & InternViT-300M   & \xmark & \cmark & \cmark & SP & I+Q          \\
PP-DocBee \shortcite{ppdocbee}         & preprint & QA          & V          & Qwen2-VL-2B    & ViT              & \xmark & \xmark & \cmark & SP & I+Q          \\
mPLUG-DocOwl2 \shortcite{mplugdocowl2} & ACL & KIE, QA     & V          & mPLUG-Owl2     & ViT              & \cmark & \xmark & \cmark & MP & I+Q          \\
TokenFD \shortcite{guan2025token} & ICCV & QA, KIE & V  & InternLM (2B, 8B) & ViT & \cmark & \cmark & \cmark & SP & I + Q\\
%    \midrule
%\multicolumn{12}{c}{\scriptsize \textit{\textbf{Note}.} KIE: Key Information Extraction; QA: Question Answering; DC: Document Classification; T: Text; L: Layout; V: Vision; MP: Multi-Page; SP: Single Page;} \\
%\multicolumn{12}{c}{\scriptsize I: Image; Q: Question; B: Bounding Box; CoT: Chain of Thought; Cap.: Captioning; Grd.: Grounding; Task: Task Information; VL: Vision-Language.} \\
\bottomrule
\end{tabular}
\end{adjustbox}
\caption{Comparison of existing MLLM-based VRDU frameworks. Mod.: Input modality; KIE: Key Information Extraction; QA: Question Answering; DC: Document Classification; T: Text; L: Layout; V: Vision; MP: Multi-Page; SP: Single Page; I: Image; Q: Question; B: Bounding Box; CoT: Chain of Thought; Cap.: Captioning; Grd.: Grounding; Task: Task Information; VL: Vision-Language.}
\label{tab:frameworks}
%\vspace{-2em}
\end{table*}}

Early VRDU frameworks relied on manually crafted rules and domain-specific heuristics \cite{watanabe1995layout,seki2007information}, which experienced a sudden performance drop on unseen documents across domains or with diverse layouts. Conventional deep learning approaches employed CNNs \cite{chargrid,yang2017learning} and RNNs \cite{bertgrid} to leverage visual or textual features, facilitating more informative representations. However, these methods typically do not effectively integrate the diverse modalities in documents, limiting their capacity to capture the rich semantic structure inherent in visually rich documents. With the success of pretraining techniques in language modeling, numerous VRDU models \cite{layoutlmv3, bros, lyu2024structextv3} have been pretrained on large-scale scanned or PDF document datasets, enabling more effective fusion of visual, textual, and layout features for robust multimodal representation. However, their effectiveness is constrained by the scope and diversity of their pretraining data, often necessitating substantial fine-tuning to achieve cross-domain generalizability. 

Recently, MLLMs \cite{gpt4o,llava}, trained on massive visual and linguistic datasets, have demonstrated powerful representational capabilities and extensive world knowledge, enabling a deeper understanding of text-dense images with diverse visual appearances and complex spatial layouts. By combining the superior text understanding of LLMs \cite{touvron2023llama} with visual encoders \cite{vit} that capture image content and layout information, MLLM-based VRDU frameworks have demonstrated strong performance across diverse document question-answering and information-extraction tasks, and generalizability across domains without task-specific fine-tuning. 

This paper provides a comprehensive survey of recent developments in MLLM-based VRDU frameworks. Previous surveys have either focused on a broad analysis of the diverse capabilities of MLLMs \cite{caffagni2024revolution} or examined techniques applied to specific document understanding tasks, such as document layout analysis \cite{binmakhashen2019document}, question answering \cite{barboule2025survey}, and relation extraction \cite{delaunay2023comprehensive}. A recent study provides \cite{ding2025deep} an overview of deep learning-based frameworks for VRDU but lacks a systematic perspective on MLLM-based approaches. In contrast, this paper provides an analysis of the MLLM-based VRDU frameworks from the aspects of \textbf{Framework Architecture} that covers both OCR- and OCR-free models (Sec~\ref{framework_architecture}), \textbf{Multimodal Representation} (Sec~\ref{input_representation}), \textbf{Training Strategies} (Sec~\ref{sec:training}), and \textbf{Inference Prompt Setting} (Sec~\ref{sec:inference}). We also include a detailed discussion of the challenges of VRDU and provide a critical analysis of the trend and future directions (Sec~\ref{future}). 
Notably, this survey is limited to methods that leverage MLLMs for document-level understanding, excluding multi-document applications, non-LLM-based methods, and MLLMs without VRD-specific adaptations.

\section{Framework Architecture}
\label{framework_architecture}

%This section reviews MLLM-based VRDU frameworks from the perspective of model architecture. 
%Depending on how multimodal features are acquired, model architectures are typically classified as either OCR-dependent or OCR-free frameworks~\footnote{See Appendix~\ref{app:model_details} to quantitative analysis and more details.}.
%\vspace{-0.7em}
\paragraph{General MLLM for VRDU.} 

Many closed- \cite{gemini1.5} and open-source \cite{internvl} general-domain MLLMs have been widely adopted for VRDU tasks and have demonstrated promising performance \footnote{Refer to Appendix~\ref{app:performance} for performance analysis.}. However, the text-dense, visually rich, and layout-sensitive nature of VRDs exposes fundamental limitations of general-domain MLLMs when applied to VRDU, including weak layout inductive bias, sensitivity to OCR noise, and hallucination on these knowledge-intensive tasks. Moreover, the wide range of downstream VRDU applications necessitates specialized techniques that adapt existing LLM backbones (as shown in Figure~\ref{tab:frameworks}) through VRDU-specific multimodal representations, training objectives, and inference paradigms. In addition, as VRDU tasks are often knowledge-intensive and safety-critical, locally tuning open-source general-domain LLMs on private document collections is essential for practical deployment in sensitive domains such as finance and industrial applications.

\paragraph{OCR-Dependent Frameworks.} 

\begin{figure}[t]
    \centering
    \includegraphics[width=\linewidth]{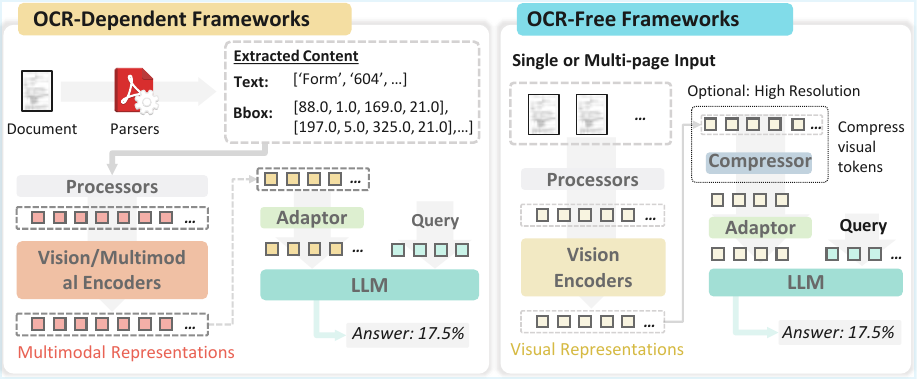}
    \caption{General OCR-dependent and OCR-free framework architectures. }
    \label{fig:architecture}
\end{figure}

As shown in Figure~\ref{fig:architecture}, OCR-dependent frameworks leverage off-the-shelf tools to extract textual and layout information from scanned or PDF documents. This extracted data, in combination with the document image, is typically fed into multimodal encoders to generate joint representations. Some models~\cite{docllm,icld3ie} input the extracted text directly into LLMs, while others~\cite{luo2024layoutllm,gpe} incorporate visual~\cite{vit} or multimodal encoders~\cite{layoutlmv3} to project those cues into language space via various adaptors or projects. These systems rely on external tools to capture structural information without extensive pretraining (e.g., text recognition). 
However, reliance on OCR or parsing tools can introduce cumulative errors, especially in handwritten or low-quality scanned documents, hindering the development of fully end-to-end models. Additionally, using low-resolution inputs may reduce the expressiveness of document representations, limiting the overall performance.

\paragraph{OCR-Free Frameworks.} 

OCR-free approaches have been introduced for end-to-end VRD understanding tasks. These frameworks bypass text extraction by directly processing document images. Visual features are extracted via one or more vision encoders, fused with the user query, and decoded by an LLM to generate responses. Representative models include Donut \cite{donut}, mPLUG-DocOwl \cite{mplugdocowl}, and UReader~\cite{ureader}.
Accurate comprehension of fine-grained text in these OCR-free settings requires high-resolution images, which, in turn, lead to lengthy visual sequences requiring visual compression modules \cite{hrvda,mplugdocowl2}. Moreover, effective text recognition in these models often relies on large-scale pretraining or instruction-tuning to integrate textual and layout features via tasks such as text spotting \cite{textmonkey} and image captioning \cite{feng2024docpedia}. This paradigm, however, demands substantial dataset construction and considerable computational resources, posing practical challenges.

\section{Multimodal Representation}
\label{input_representation}

%Unlike plain text or natural scene images, VRDs contain dense textual content closely aligned with complex visual and layout elements. %Effective VRD understanding requires encoding this multimodal information to support tasks like information extraction and knowledge-intensive question answering. 
%This section reviews MLLM-based feature (text, vision, and layout) encoding and fusion methods for the VRD domain (see Figure~\ref{fig:feature_representation}).
\begin{figure*}[t]
    \centering
 \includegraphics[width=\linewidth]{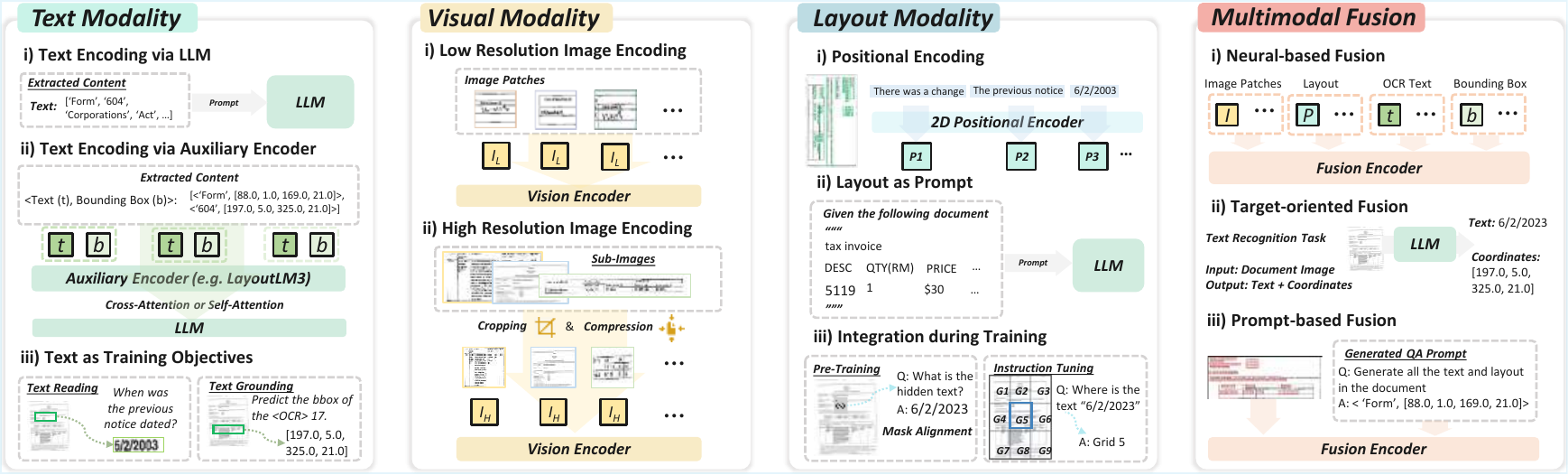}
    \caption{Multimodal feature representation and fusion mechanisms. }
    \label{fig:feature_representation}
    %\vspace{-1.5em}
\end{figure*}

\subsection{Text Modality}
\label{sec:text_modality}
%As a core modality in VRDU tasks, textual information serves both to represent document content and to deliver prompts or instructions that guide LLM-based answer generation. This section focuses on methods for document content representation, while prompting strategies are discussed in Section~\ref{sec:inference}. 
%Textual information is essential for capturing the core content, semantics, and context of documents. 
OCR-dependent methods rely on external tools to extract text for encoding, while OCR-free models use document images directly.
%Since VRDs are typically formatted as PDFs or images, OCR-dependent frameworks commonly employ standard OCR and PDF parsing tools to extract text and layout features. The extracted content is subsequently processed using tokenizers or processors associated with the chosen framework. The text encoding methods can be categorized into two ways.

\paragraph{Text Encoding via LLM. }
Given the frequent text recognition challenges faced by MLLMs, stemming from low-resolution inputs or undertrained vision encoders, off-the-shelf OCR-extracted text is commonly embedded directly into LLM prompts to enhance document comprehension~\cite{docllm,dockd} (see Figure~\ref{fig:feature_representation}). However, the extracted content is often unordered; to address this, frameworks such as ICL-D3IE~\cite{icld3ie} and LLaVA-Read~\cite{llavaread} employ the XY-cut algorithm to reorder the text sequence. Additionally, to handle long documents, some methods segment the text into chunks, though this may introduce semantic discontinuities~\cite{pdfwukong}. 
In sum, directly adding extracted text to prompts improves context and reduces reliance on additional encoders; however, performance remains limited by OCR and LLM errors.

\paragraph{Text Encoding via Auxiliary Encoder.}
To enhance multimodal integration, many frameworks introduce auxiliary encoders to enhance text embeddings. Several methods \cite{luo2024layoutllm,gpe} enhance text representation and multimodal fusion by feeding extracted text, image patches, and bounding boxes into pretrained LayoutLMv3~\cite{layoutlmv3}. Notably, Zhu et al.~\shortcite{gpe} propose a ROI Aggregation module that aggregates fine-grained tokens (e.g., words) into object-level features (e.g., paragraphs), facilitating downstream object-level contrastive learning. Instruct-Doc \cite{instructdoc} introduces an enhanced Q-Former \cite{blip2}, termed \textit{Document Former}, serving as a bridging module that integrates visual, textual, and layout information into the LLM input space via cross- and self-attention. In sum, external encoders improve representations but require additional pretraining and fine-tuning to align with LLMs' latent spaces. % While such external encoders improve textual representations via multimodal fusing and pretrained knowledge, they necessitate additional pretraining and instruction tuning to align the encoder outputs with the LLM input space.

\paragraph{Text as Training Objectives.}
Some frameworks rely exclusively on document images as input to predict answers. Models such as mPLUG-DocOwl~\cite{mplugdocowl} and LLaVA-R~\cite{llavar}, built upon mPLUG-Owl~\cite{mplugowl}, demonstrate strong OCR capabilities and are further instruction-tuned on diverse VRDU benchmarks. Other approaches incorporate text recognition, detection, and spotting tasks~\cite{wang2023towards,unidoc} to integrate text information. To better understand the hierarchical structure of documents, \citet{mplugdocowl15,mplugdocowl2} propose a multi-grained text localization task spanning the word-to-block level.
While these methods deliver robust results using only visual inputs, they place heavy demands on pretraining and fine-tuning. Additionally, high-resolution images are often necessary to accommodate extremely long visual sequences and to preserve fine-grained features~\cite{hrvda,texthawk}.

\subsection{Visual Modality}
To integrate visual information, OCR-dependent frameworks use extracted text and coarse visual cues, thereby enabling the use of \textbf{lower-resolution} images. In contrast, OCR-free frameworks require direct text recognition, demanding fine-grained perception and \textbf{high-resolution} inputs. See the Appendix~\ref{app:framework_detail} for input resolution details.

\paragraph{Low Resolution Image Encoding.}
\label{sec:visual_modality}
Some frameworks directly feed image patches into pretrained vision encoders to obtain patch embeddings \cite{pdfwukong,instructdoc}. Others \cite{mdocagent,luo2024layoutllm,doclayllm} employ pretrained VRDU models, i.e., LayoutLMv3~\cite{layoutlmv3}, to extract multimodal-enhanced visual embeddings. Due to the limitations of low-resolution inputs in capturing fine-grained details, recent works have adopted dual-encoder architectures that process both low- and medium-resolution images~\cite{ureader,llavaread}, followed by visual feature compression techniques to manage the increased feature volume. While using low-resolution images offers a straightforward pathway to multimodal understanding, achieving effective alignment often requires additional pretraining and instruction tuning. Moreover, the absence of fine-grained visual detail often necessitates additional OCR tools to extract text for accurate VRD interpretation. 

\paragraph{High Resolution Image Encoding.}
To capture fine-grained level information for end-to-end training and inference, many frameworks support high-resolution image input. For ViT-style \cite{vit} pretrained vision encoders, \citet{mplugdocowl15} splits high-resolution images into predefined sub-images. To handle images of various shapes, UReader \cite{ureader} introduces a \textit{Shape-Adaptive Cropping Module} that adaptively divides images into fixed-size sub-images using grids of various shapes. %This technique is now widely adopted \cite{mplugdocowl15,mplugdocowl2,texthawk,texthawk2}. 
However, the image cropping may disrupt semantic continuity across sub-images. To address this, \citet{textmonkey} introduced a \textit{Shifted Window Attention} to enhance cross-sub-images connection via self-attention. 
In short, high-resolution images support fine-grained information extraction, but efficiently processing the resulting large number of visual tokens remains challenging, requiring a balance between resource usage and the number of visual tokens. 

\paragraph{Visual Feature Compression.}
%Visual Feature Compression techniques are employed Tto mitigate the large number of visual tokens generated by high-resolution images, . 
\citet{texthawk, texthawk2} utilize Q-Former \cite{blip2}, while \citet{textmonkey} adopts the \textit{Resampler} from Qwen-VL \cite{Qwen2-VL} to reduce the number of visual tokens. Considering the layout-aware nature of VRDs, \citet{mplugdocowl15} introduces a convolutional module that preserves layout by compressing horizontal features and reducing the number of tokens. It further enhances this with layout-aware cross-attention to handle multi-page input. \citet{hrvda} use a \textit{Content Detector} to filter non-informative tokens by segmenting text-rich regions, while \citet{dockylin} propose eliminating low-information areas and clustering and aggregating the remaining features.

%High-resolution images enable fine-grained information extraction and support end-to-end inference. However, efficiently processing the resulting large number of visual features remains challenging, as it requires balancing resource usage with preservation of semantic continuity and layout structure. While OCR-free frameworks eliminate the need for OCR during inference, many still rely on OCR-generated annotations during pretraining, inheriting limitations from OCR inaccuracies. Furthermore, existing methods often overlook semantic and logical connections within documents, such as references from text to tables, particularly in multi-page scenarios.

\subsection{Layout Modality}
\label{sec:layout_modality}

Unlike natural scene images, VRDs feature dense text and complex layout structures. Methods for encoding layout information can be categorized into positional encoding-based, prompt-based, and task-oriented approaches.  %To enable structure-awareness, some frameworks refer to pretrained VRDU models \cite{layoutlm,lilt} to design more effective Positional Encodings to incorporate layout information. 
%Recent methods enhance encoding of layout information by integrating layout features into prompts and introducing layout-aware pretraining and instruction-tuning.

\paragraph{Positional Encoding.}
OCR-dependent models use OCR tools to extract textual and layout information, combining text embeddings with 2D positional encodings \cite{layoutlm} to incorporate layout into LLMs \cite{mdocagent,instructdoc}. However, these approaches require extra training for feature alignment. In contrast, \citet{gpe} assigns unique positional embeddings to attention heads based on multi-dimensional layout features without altering the model architecture or requiring further pretraining. 
\citet{docllm} treats layout as a separate modality and introduces disentangled spatial attention for cross-modal interactions without visual encoders. 
\citet{laytokenllm} addresses long-context inference limits by encoding layout as a single token sharing the position with its text. However, these methods implicitly integrate layout information and rely heavily on large-scale pretraining, resulting in high computational costs and reduced effectiveness for tasks that demand explicit layout understanding.

\paragraph{Layout as Prompt.}
To integrate explicit layout information, some frameworks include layout details in prompts alongside the user query and document content. \citet{icld3ie} introduces an in-context learning based approach to incorporate layout-aware demonstrations into bounding box representations. \citet{lapdoc} and \citet{lmdx} encode layout into text sequence through rule-based verbalization or quantized coordinate tokens. These methods enable layout-awareness without training. However, these methods increase input length, rely on LLMs to interpret layout as text, and overlook visual cues essential for encoding relative positional information.

\paragraph{Integrating During Training.}
OCR-free frameworks incorporate text by formulating recognition and detection tasks that also aid in understanding layout  \cite{wang2023towards,unidoc}. To further enhance this, some models \cite{marten,llavaread} leverage layout-aware pretraining tasks (Section~\ref{sec:pretraining}) and layout-specific instruction-tuning tasks, such as visual grounding~\cite{hrvda,textmonkey} and table reconstruction~\cite{doclayllm}. However, these methods typically require large-scale datasets for pretraining or instruction tuning, leading to substantial computational costs and data bottlenecks.

\subsection{Multimodal Fusion}
\label{sec:multimddal_fusion}
We categorize multimodal fusion methods into four types: direct, neural-based, task-oriented, and prompt-based. Direct fusion relies on simple feature summation or concatenation with alignment training, while this survey primarily focuses on the latter three approaches.

\paragraph{Neural-based Fusion.}
The simplest multimodal feature encoding uses external document encoders such as LayoutLMv3 \cite{layoutlmv2}, which fuse multimodal features via self- or cross-attention and leverage pretraining knowledge. \citet{docllm} stands out by employing a layout-aware transformer with disentangled attention over text and spatial layouts, enabling effective document understanding without requiring image encoders. In OCR-free frameworks, visual encoders extract visual cues, with adaptors like LoRA \cite{texthawk2} or linear projectors \cite{llavar,wang2023towards} mapping features into the language space. \citet{masryalignvlm} propose a method that maps visual features to a weighted textual embedding to reduce misalignment issues observed in previous approaches. These neural-based fusion methods benefit from dedicated encoders or modified architectures, but often require extensive pretraining or SFT and face challenges in scalability, computational overhead, and adaptability to diverse document layouts.

\paragraph{Target-oriented Fusion.}
%We have summarized target-oriented feature encoding approaches, such as OCR \cite{texthawk2} and visual grounding \cite{mplugdocowl15}, which facilitate multimodal fusion. 
Target-oriented strategies establish multimodal connections through supervised objectives that span the input-to-output space \cite{mplugdocowl15} and are widely applied to text and layout features in OCR-free frameworks. For instance, in text recognition tasks, models are trained to map visual features directly to text and spatial coordinates, thereby aligning fusion with task-specific goals. While these approaches improve end-to-end multimodal integration, they also increase demands on data preparation, annotation quality, and training complexity in practice.

\paragraph{Prompt-based Fusion. }
Prompts for multimodal tasks may include text, images, and bounding box coordinates. While many frameworks adopt Layout-as-Prompt strategies to encode layout information, others use Chain-of-Thought (CoT) reasoning to further enhance multimodal learning. For example, \citet{luo2024layoutllm} utilizes a \textit{LayoutCoT} approach that divides reasoning into question analysis, region localization, and answer generation, explicitly modeling spatial layout. \citet{doclayllm} leverages CoT pretraining and CoT annealing to support layout-aware reasoning for VRDU. However, these methods often depend on predefined reasoning strategies, intermediate-step evaluations, and well-trained prior frameworks, limiting their generalizability. 

\section{Training Paradigms}
\label{sec:training}

\begin{figure*}[t]
    \centering
 \includegraphics[width=\linewidth]{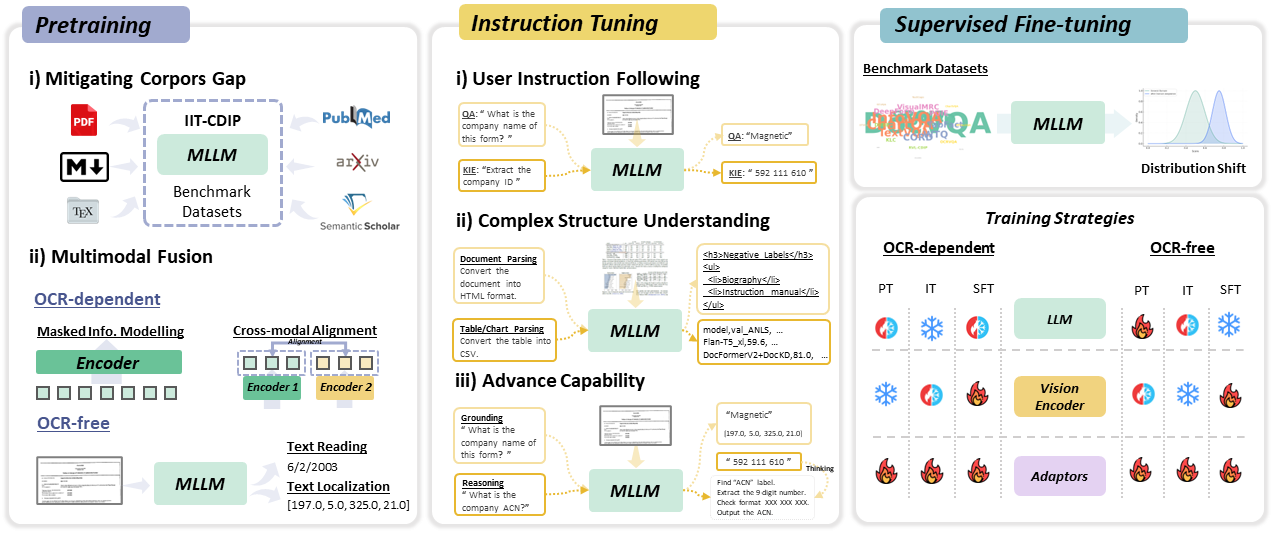}
    \caption{MLLM-based VRDU framework training paradigms.}
    \label{fig:training_stratigies}
\end{figure*}

To facilitate multimodal understanding, instruction following, and domain adaptation, various training tasks and strategies have been developed, as illustrated by Figure~\ref{fig:training_stratigies}.
%Finally, the overall training workflow is outlined, highlighting the adopted strategies and indicating which components are kept frozen.
% \clearpage 

\subsection{Pretraining Strategies} 
\label{sec:pretraining}
To enhance mono- and multi-modal document understanding, VRDU frameworks adopt various self-supervised pretraining tasks, such as masked information modeling and cross-modality alignment~\cite{ding2025deep}. OCR-dependent frameworks typically utilize pretrained VRDU models or vision encoders to obtain enriched multimodal representations. Some models propose additional self-supervised learning tasks (e.g., \citet{li2024enhancing} applies object-level contrastive learning between visual and multimodal features). 
\citet{docllm} introduces a transformer architecture with disentangled spatial-text attention to perform block-wise text infilling to enhance text-layout correlation modeling. OCR-free frameworks~\cite{llavaread,mplugdocowl15} focus on pretraining tasks like text recognition, detection, and captioning to integrate text and layout information. \citet{mplugdocowl2} further targets multi-page layout coherence. \citet{feng2024docpedia} aligns frequency features with LLMs through text-centric pretraining. Although these self-supervised tasks are effective in fusing multimodal features and learning general knowledge, they remain computationally intensive and often lack instruction-based tuning, limiting their capacity to follow real-world user instructions.
\vspace{-0.5em}
\subsection{Instruction Tuning} 
To benefit task orientation in LLM-based frameworks, many VRD approaches, following InstructGPT~\cite{instructgpt}, are trained on instruction-response pairs to better align model outputs with user prompts. Pretraining tasks such as text reading, recognition, and image captioning are reformulated as instruction-based formats, with images paired with task descriptions. Beyond improving multimodal fusion, goal-oriented tasks, including VRD question answering \cite{pdfmvqa}, key information extraction \cite{formnlu}, and VRD classification \cite{rvlcdip}, are conducted on large-scale datasets. For better generalizability, some frameworks synthetically generate large instruction-tuning datasets (See Appendix~\ref{app:dataset} for more details). To further improve localization and information extraction, \citet{wang2023towards} and \citet{unidoc} propose predicting answers alongside bounding boxes, thereby enhancing the framework's reliability. Instruction tuning not only strengthens user query understanding but also boosts multimodal fusion. Instruction tuning on large-scale datasets substantially enhances zero-shot performance. However, the requirement for extensive training data leads to substantial resource consumption. Furthermore, synthetic datasets, often generated with off-the-shelf OCR tools and LLMs, may yield low-quality QA pairs, particularly in low-resource domains such as scanned documents, thereby impacting zero-shot performance. 

%\subsection{Supervised-Fine Tuning} Pretraining and instruction-tuning improve the generalizability and understand user prompt properly. However, many frameworks 

\subsection{Training Strategies} MLLM-based document understanding frameworks typically consist of multiple sub-modules to encode multimodal information and are trained in a stepwise manner. Few frameworks leverage in-context learning \cite{icld3ie} or multimodal prompts \cite{lmdx} to develop training-free architectures. The majority, however, involve pretraining to capture general-domain knowledge, followed by instruction tuning to improve the interpretation of user prompts. Furthermore, some frameworks are subsequently \textbf{Supervised Fine-Tuned} on benchmark datasets \cite{docllm,gpe} or a synthetic set \cite{dockd} to enhance domain-specific adaptation.
To integrate multimodal information, these frameworks mainly employ an LLM with various multimodal encoders \cite{mdocagent,pdfwukong}, sometimes incorporating adaptors \cite{mplugdocowl15,lu2024bounding} or linear projectors \cite{park2024hierarchical} for fusion or alignment. Depending on the training stage, sub-modules may be either trainable or frozen, balancing the acquisition of new knowledge with the preservation of valuable information from the original backbone. 
%The configurations of each sub-module are summarized as follows:

\paragraph{LLM Backbone.} As most LLMs are extensively pretrained on large-scale datasets and capture broad knowledge, many frameworks freeze the LLM, using it solely to generate human-understandable outputs. In frameworks involving pretraining or instruction tuning \cite{dockylin,hrvda}, freezing the LLM backbone helps preserve its knowledge and reduce training costs. However, some approaches enable LLMs to be trained during continued pretraining \cite{laytokenllm}  or instruction tuning \cite{doclayllm} to better capture VRD domain knowledge and enhance multimodal alignment. In supervised fine-tuning stages, the LLM backbone is typically made trainable to adapt to the target domain \cite{llavar}.

\paragraph{Multimodal Encoders.} They are employed to encode multimodal features, which are subsequently aligned with LLM text representations by projectors or adapters. Similar to LLM backbones, vision \cite{vit}, and multimodal encoders \cite{layoutlmv3} are often kept frozen during pretraining to preserve learned knowledge \cite{texthawk2,llavar}. Feng et al.~\shortcite{feng2024docpedia} use a Swin Transformer to encode frequency-domain images, pretrained from scratch. To enhance multimodal feature learning, Li et al.~\shortcite{li2024enhancing} make the ViT encoder trainable while freezing LayoutLMv3, enabling knowledge distillation via contrastive learning. During instruction tuning, vision encoders are typically unfrozen to improve alignment and task-specific adaptation \cite{dockylin,hrvda}. Conversely, in dual-encoder frameworks, vision encoders with inputs at diverse resolutions are often frozen to enhance the representation of hierarchical inputs. In supervised fine-tuning, there is no standard practice for encoder trainability.

\paragraph{Projectors and Adaptors.} 
They play a crucial role in feature alignment and lightweight tuning. Projectors are typically employed to align visual or layout features with the LLM input space \cite{park2024hierarchical} and encode layout information \cite{instructdoc}. These modules are mainly trainable throughout the entire training process. Adaptors, on the other hand, are designed for efficient, task-specific tuning, often leveraging LoRA-style updates \cite{mplugdocowl,mplugdocowl15} or cross-attention mechanisms \cite{textmonkey,texthawk} to integrate multi-aspect inputs with minimal parameter changes. Plug-and-play components, such as visual abstractors \cite{mplugdocowl} or compressors \cite{mplugdocowl2}, have also been introduced to reduce the dimensionality of visual features. These adaptors are usually trained during instruction tuning or during supervised fine-tuning. 

% \subsection{Training Datasets Overview}
% % One paragraph to summarise the trend e.g data usage for pretraining, instruction tuning and supervsied fine-tuning. 
% Diverse datasets are required to meet specific training objectives. Pretraining typically leverages large-scale cross-domain VRD collections to reduce domain gaps and enhance multimodal fusion, sometimes requiring more domain-specific data (e.g., medical, slides) to improve domain awareness. For instruction tuning, synthetic datasets are often used to strengthen instruction-following and reasoning abilities, particularly in OCR-free frameworks that generate instruction-aligned OCR or layout understanding tasks. Additionally, SFT is commonly applied on original or post-processed benchmark datasets (e.g., converting key-value pairs into QA format) to further boost performance. For more dataset details, see Appendix~\ref{app:benchmark_dataset}.

\section{Datasets}
\subsection{Pretraining Datasets.} 

The goal of pretraining is to enhance multimodal understanding and improve generalization across VRDU tasks. MLLM-based approaches commonly perform continued pretraining on large-scale, cross-domain document collections such as IIT-CDIP~\cite{iitcdip}, which contains over 6 million scanned documents across diverse domains, though lacking explicit layout annotations, often supplemented with OCR-derived bounding boxes. RVL-CDIP~\cite{rvlcdip}, a curated subset with 400,000 documents across 16 categories, is widely used for document classification. Beyond these general-purpose datasets, recent frameworks~\cite{llavar, wang2023towards} have introduced self-collected datasets to target domain-specific or task-oriented scenarios, including slide decks~\cite{feng2024docpedia}, academic papers~\cite{docllm}, and other structured document types~\cite{texthawk2}.

\subsection{Instruction-tuning Datasets.} 

Instruction-tuning aims to enhance a model's understanding of user queries. 
Many frameworks~\cite{zhangtoken,park2024hierarchical} perform instruction-tuning directly on benchmark document collections to improve downstream task performance. Others~\cite{luo2024layoutllm,hrvda} generate large-scale synthetic datasets using OCR tools to extract text and layout information from VRD-related benchmarks such as layout analysis~\cite{publaynet} and document classification~\cite{rvlcdip}. Instruction-response pairs are then created based on predefined task definitions. Some frameworks also construct their own multi-domain datasets to improve generalizability and prevent data leakage~\cite{vary,unidoc}. Instruction-tuning is critical for domain adaptation and accurate instruction interpretation. As shown by Table~\ref{tab:instruction_tuning_dataset}, some frameworks increasingly generate synthetic instruction-tuning datasets tailored to their architectures, prioritizing alignment over generalizability achieved through benchmark-based tuning. 
\subsection{Benchmark Datasets}
\label{app:benchmark_dataset}
%Based on differences in downstream tasks and the benchmark dataset's domain, we list the widely used VRDU dataset and its key attributes in Table~\ref{tab:vrd_benchmark_dataset}, including both VRD-related Key Information Extraction (KIE) and Visual Question Answering (VQA). 

\paragraph{Key Information Extraction}
Benchmarks for Key Information Extraction~(KIE) are shifting from early schema-constrained tasks (e.g., SROIE~\cite{sroie}, FUNSD~\cite{funsd}) toward larger, multilingual, cross-domain, multi-page, and open-vocabulary challenges. While form-like structures (e.g., DocILE~\cite{docile}, Form-NLU~\cite{formnlu}) still dominate the landscape, modern resources such as KVP10k~\cite{kvp10k} and CC-OCR-KIE~\cite{cc-ocr} focus on \textit{open-category} extraction without predefined schemas.
Furthermore, a clear trend of dataset consolidation and multilingual expansion has emerged.

\paragraph{Visual Question Answering.} has undergone a comparable evolution, shifting from early single-page, text-centric retrieval to benchmarks that probe multiple dimensions of complexity. 
This progression is reflected in broader multilingual coverage (e.g., MTVQA~\cite{MTVQA}, JDocQA~\cite{JDocQA}) and more diverse, multi-domain settings (e.g., DUDE~\cite{dude}). Recent datasets increasingly emphasize long-context comprehension over multi-page documents: benchmarks such as LongDocURL~\cite{longdocurl}, BRIDGE~\cite{xiang2026bridge} and MMLongBench-Doc~\cite{MMLongBench} contain documents averaging dozens of pages and often demand non-trivial cross-page evidence aggregation and reasoning. In parallel, reasoning requirements have deepened toward domain-specific expertise, as illustrated by vision-essential physics problem solving in SEEPHYS~\cite{SeePhys}. Finally, dataset scale has expanded substantially, reaching millions of instances in collections such as MMVQA~\cite{mmvqa}, thereby enabling rigorous stress-testing of the capacity and reasoning limits of modern multimodal models.

%\paragraph{Other Domain Datasets} Many frameworks are evaluated on other domain-specific datasets as well, including those for chart understanding and webpage analysis. For instance, InfoVQA \cite{infovqa2022} focuses on visual question answering for information-centric records. Benchmarks like WTQ \cite{pasupat2015wtq} and TabFact \cite{tabfact2020} assess a model's ability to reason over tabular data, and ChartQA evaluates chart comprehension skills. Additionally, TextVQA \cite{textvqa2019} and TextCaps \cite{textcaps2020} target text recognition and semantic reasoning in natural images. 

\section{Inference Prompt Setting}\label{inference}
\label{sec:inference}
MLLM-based frameworks adopt diverse prompt formats depending on their architecture. For OCR-free frameworks in Table~\ref{tab:frameworks}, the prompt typically includes a document image, occasionally multiple pages \cite{mplugdocowl2,marten}, alongside a textual user query. Some frameworks not only predict answers to user queries but also localize bounding boxes, often requiring an additional prompt for localization \cite{wang2023towards,unidoc}. OCR-dependent frameworks first preprocess input using off-the-shelf tools to extract textual and layout information. Vision-free models \cite{icld3ie,docllm} process only the extracted content alongside the query. In contrast, vision-dependent models also incorporate the document image into the vision \cite{pdfwukong} or into multimodal encoders \cite{doclayllm}, aligning visual and textual features for the final prediction. Furthermore, some frameworks integrate layout information into prompts via bounding boxes \cite{gpe} or markdown-style formatting. The inference strategies are closely tied to the model architecture and reflect a growing trend toward unified, multimodal understanding and layout-aware reasoning to improve document comprehension accuracy and versatility.

\section{Challenges and Future Direction}\label{future}
%\subsection{Reinforcement Learning}
%In this survey, we provided a comprehensive review of recent MLLM-based frameworks for VRDU, systematically examining model architectures, multimodal representations, training, and inference strategies. Despite substantial advances in this field, several critical challenges and research directions remain largely underexplored.

\paragraph{Synthetic Data.}
Acquiring high-quality, manually curated datasets for new document collections is often quite costly. Leveraging synthetically generated datasets offers a cost-effective alternative for adapting to the target domain ~\cite{ding2025syndoc,ding2026docs2synth}. For large-scale instruction-tuning, many frameworks generate instruction-response pairs using benchmarks, templates, or LLMs. However, these synthetic datasets often lack validation, resulting in noise. 
%Verification, particularly with LLMs as evaluators, is crucial to ensure data reliability. 
Since synthetic data may not fully capture real user input, future research should prioritize human-in-the-loop and reinforcement learning approaches to improve authenticity and task relevance.

\paragraph{Long Document Understanding.} 

In practice, VRDs frequently span multiple pages; however, most existing frameworks are tailored for single-page inputs. Multi-page approaches typically rely on retrievers to identify relevant pages, which are then processed by MLLM-based VRDU systems. These methods often fall short of capturing semantic and logical dependencies among document entities, resulting in incomplete contextual understanding. 
%Hu et al.~\cite{mplugdocowl2} introduce multi-page scale pretraining to improve cross-page comprehension, though their focus remains primarily on text recognition rather than semantic and logical relations. 
Furthermore, handling long input sequences remains challenging, as existing multi-page benchmarks focus mainly on extractive tasks and rarely support complex multi-hop or multimodal reasoning.

\paragraph{Multilingual VRDU.} 

Most existing models and benchmarks remain heavily English-centric, limiting their generalization to documents with diverse languages and layouts. This bias is further amplified by large-scale pretraining corpora that predominantly reflect English document structures, leading to performance degradation in low-resource settings. Although few multilingual datasets have been proposed \cite{xu2022xfund,chen2025mosaicdoc}, future research should explore more multilingual and culturally diverse benchmarks, language-agnostic representation learning, and hybrid approaches to mitigate linguistic bias to handle real-world document diversity.

\paragraph{Effective RAG Framework.} 

While RAG has become a common paradigm \cite{jain-etal-2025-simpledoc,zhang2026stindex, faysse2025colpali}, existing approaches often exhibit brittle retrieval due to layout ambiguity and misaligned multimodal embeddings, leading to unreliable evidence selection. Moreover, most RAG pipelines decouple retrieval from reasoning and remain largely text-centric, limiting their ability to capture spatial and visual semantics in complex documents. Future work should explore multimodal RAG frameworks that support iterative reasoning and dynamic evidence refinement, and enable more robust and interpretable VRDU.

\paragraph{Agentic LLM in VRDU.}

Recent works \cite{mdocagent, sun2025docagent} incorporate external tools (e.g., PDF parsers or retrievers) to generate intermediate outputs, enhancing both the accuracy and interpretability of practical VRDU applications. However, future research should explore a wider variety of agent types and architectural innovations to enable automatic handling of diverse formats, cross-domain scenarios, and fine-grained elements such as charts and tables. Additionally, challenges in agentic AI, such as multi-agent coordination and knowledge conflicts, remain significant barriers to broader adoption for VRDU.

%\noindent\textbf{Scaling Law and Domain Adaptation. }
%Most frameworks utilize large-scale datasets for pretraining or instruction-tuning, acquiring general knowledge that boosts performance in zero-shot and fine-tuned settings. Nonetheless, a significant performance gap remains compared to fine-tuned BERT-style VRDU models on domain-specific tasks. While scaling laws suggest performance improves with larger models and data, gains diminish under domain shifts or distributional discrepancies. Despite strong generalizability, large models often falter in domain adaptation due to reliance on heterogeneous corpora that lack fine-grained semantics and layout cues. Their substantial computational and data demands further hinder adaptation in low-resource or specialized domains with limited annotated data. Incorporating diverse agents and synthetic datasets may offer a promising direction to enhance domain adaptation.

\newpage
\section*{Limitations}
While this survey offers a comprehensive overview of MLLM-based VRDU research, our analysis is necessarily qualitative. It does not provide exhaustive head-to-head comparisons, as the field's rapid evolution and breadth prioritize trend summarization over detailed benchmarking. Although academic advances are thoroughly reviewed, discussion of real-world deployments and industrial challenges remains limited, in part because many practical applications are proprietary and unpublished. In future work, we aim to provide more quantitative meta-analyses, incorporate insights from industrial adoption, and continuously update the survey to capture the latest developments as the field progresses.

\section*{Acknowledgements}
This research was supported by the Australian Research Council (ARC) Training Centre for Critical Resources for the Future (CCRF) under grant number IC230100035. This work was also supported by the National Library of Medicine [grant numbers R01LM014344, R01LM014573] and the National Science Foundation (NSF) [grant numbers 2145640, 2139899].

\bibliography{custom}

\appendix

\newpage
\clearpage
\section{More Framework Details}
\label{app:model_details}

%\subsection{Framework Key Contributions}
%Table~\ref{tab:key_techniques_mllm_vrdu} presents a comparative summary of key techniques underlying recent MLLM-based frameworks for VRDU. Each model employs distinct innovations to enhance multimodal representation and reasoning, spanning novel positional encoding schemes, advanced visual token compression, multi-agent architectures, and instruction-tuning strategies tailored to document tasks such as KIE, QA, and visual grounding. The table highlights how frameworks leverage document structure, spatial layout, and semantic cues through both OCR-dependent and OCR-free paradigms. By distilling the core technical contributions of each approach, this summary provides a comprehensive reference for researchers and practitioners seeking to understand the landscape and evolution of state-of-the-art MLLM-based document understanding solutions.

\subsection{Open-source Frameworks}
\begin{table*}[!t]
\centering
\small
\begin{tabular}{lll}
\toprule
\textbf{Framework} & \textbf{Model Name} & \textbf{Official Open Source Link} \\
\midrule
mPLUG-DocOwl~1.5 & DocOwl~1.5 & \href{https://github.com/X-PLUG/mPLUG-DocOwl/tree/main/DocOwl1.5}{github.com/X-PLUG/mPLUG-DocOwl/tree/main/DocOwl1.5} \\
mPLUG-DocOwl~2   & DocOwl~2   & \href{https://github.com/X-PLUG/mPLUG-DocOwl/tree/main/DocOwl2}{github.com/X-PLUG/mPLUG-DocOwl/tree/main/DocOwl2} \\
UReader          & UReader    & \href{https://github.com/X-PLUG/mPLUG-DocOwl/tree/main/UReader}{github.com/X-PLUG/mPLUG-DocOwl/tree/main/UReader} \\
KOSMOS-2.5       & KOSMOS-2.5 / 2.5-CHAT & \href{https://aka.ms/kosmos25}{aka.ms/kosmos25} \\
LLaVAR           & LLaVAR     & \href{https://github.com/SALT-NLP/LLaVAR}{github.com/SALT-NLP/LLaVAR} \\
Marten           & Marten     & \href{https://github.com/PriNing/Marten}{github.com/PriNing/Marten} \\
LEOPARD          & LEOPARD    & \href{https://github.com/Jill0001/Leopard}{github.com/Jill0001/Leopard} \\
\bottomrule
\end{tabular}
\caption{Official open-source links for some VRDU/MLLM frameworks.}
\label{tab:open_source_frameworks}
\end{table*}
Table~\ref{tab:open_source_frameworks} presents official open-source links for VRDU and MLLM frameworks, underscoring the vital role of open access in fostering transparency, reproducibility, and accelerated innovation within the research community.

\subsection{Model Training Paradigm Comparison}

Table~\ref{tab:mlmm-vrdu-training} provides a comprehensive comparison of MLLM-based VRDU frameworks across three major training stages: Pretraining (PT), Instruction-tuning (IT), and Supervised Fine-tuning (SFT). OCR-dependent models generally rely on external text extraction and have limited pretraining because they are trained on OCR-processed inputs. In contrast, OCR-free models, which operate directly on document images, demonstrate richer instruction-tuning and fine-tuning strategies, often involving frozen or LoRA-based vision and language encoders. This highlights the diverse training paradigms and modular designs adopted to balance efficiency, adaptability, and performance across frameworks.

\subsection{Model Component Details}
\label{app:framework_detail}

Table~\ref{tab:mlmm-vrdu-backbones} presents a comprehensive comparison of component configurations adopted by recent MLLM-based frameworks for VRDU, spanning both OCR-Dependent and OCR-Free paradigms. For each model, we summarize its LLM backbone (e.g., Vicuna, Qwen, LLaMA, GPT), vision encoder (e.g., CLIP, ViT, Swin), input resolution (including dynamic scaling and cropping), and specialized adaptors or projectors (e.g., LoRA, MLP, QPN) used for multimodal fusion. OCR-Dependent models typically incorporate layout-aware encoders (e.g., LayoutLMv3, DocFormer) and rely on structured textual inputs. In contrast, OCR-Free models process raw document images directly, often requiring higher resolutions and additional modules such as resamplers, visual abstractors, or cropping strategies. The table also lists the maximum supported image resolution, indicating each model's capacity for fine-grained visual understanding. This comparison highlights the increasing diversity in MLLM architectures and the adoption of lightweight tuning techniques for scalable VRDU.
\subsection{Document Parsing Tools}
\label{app:parsing_tools}
Table \ref{tab:ocr_llm_comparison} provides a comparative overview of representative OCR engines, document parsing APIs, and vision–language models for document understanding. The table highlights clear trade-offs across deployment modes, pricing models, and functional capabilities: traditional OCR engines are predominantly open-source and locally deployable but offer limited support for structured document parsing, while commercial document APIs and vision LLMs more frequently provide GPU acceleration and native document-structure extraction at the cost of cloud dependency and usage-based pricing. Recent vision–language models bridge OCR and higher-level reasoning by supporting multimodal inputs (image and PDF) and multilingual processing, yet vary substantially in openness and deployment flexibility. Overall, the comparison illustrates the evolving landscape from text-centric OCR toward multimodal, structure-aware document understanding systems.
\section{Dataset Overview.}
\label{app:dataset}
Tables~\ref{tab:pretraining_datasets} and \ref{tab:instruction_tuning_dataset} summarize the datasets used across different training stages. Pretraining typically relies on large-scale, cross-domain document corpora (e.g., IIT-CDIP, RVL-CDIP) to build general multimodal understanding, sometimes extended with domain-specific collections. Instruction-tuning datasets are constructed either from benchmark datasets or via synthetic generation to improve instruction following and domain adaptation. For downstream optimization, supervised fine-tuning commonly leverages QA-style benchmarks (e.g., DocVQA, MPDocVQA) and reformulates key information extraction datasets (e.g., FUNSD, CORD). These tables provide a structured overview of dataset sources and their roles in the training pipeline.
%Detailed descriptions of these benchmark datasets are found in Appendix~\ref{app:benchmark_dataset}. 
%While fine-tuning on these datasets can significantly boost performance, it often depends on curated training sets, which are challenging to obtain in real-world applications.

%\subsection{Pretraining Dataset}
%Similar to pretrained VRDU frameworks, MLLM-based approaches also conduct continued pretraining on large-scale document collections to improve multimodal understanding. Commonly used datasets include IIT-CDIP \cite{iitcdip}, which contains over 6 million scanned documents spanning diverse domains but lacks layout annotations, often supplemented with OCR-derived bounding boxes. RVL-CDIP \cite{rvlcdip} is a smaller, curated subset of 400,000 documents across 16 categories, widely used for classification and low-resource pretraining. Except for the two general-domain document collections, some works use self-collected document collections to enhance specific domains or enrich document collection types, as summarised in Table~\ref{tab:pretraining_datasets}.

\section{Quantitive Analysis}
\label{app:performance}
\subsection{Performance on Single Page Benchmarks}
Table~\ref{tab:dataset_comparison} highlights clear trends in the performance of general-domain LLMs/MLLM and OCR-dependent and OCR-free document understanding frameworks across several popular benchmarks. Generally, OCR-dependent models achieve consistently strong results on classic form and receipt datasets such as FUNSD, CORD, and SROIE—often exceeding 80\% accuracy, with top models such as PDF-WuKong, GPE, and DocLayLLM achieving state-of-the-art performance. In contrast, OCR-free frameworks, while demonstrating rapid progress, still lag on these traditional datasets but show remarkable advances on more visually and semantically complex benchmarks such as DocVQA, ChartVQA, and InfoVQA. Notably, the latest OCR-free models, including Texthawk2, Marten, and PP-DocBee, have begun to outperform or match OCR-dependent methods on DocVQA and chart-centric tasks, signaling a narrowing of the gap in real-world document reasoning capabilities. However, coverage remains uneven, with many OCR-free models performing poorly on specific datasets, indicating ongoing challenges with generalizability and benchmark saturation. Overall, while OCR-dependent methods remain dominant for structured text extraction, OCR-free approaches are quickly maturing and expanding the frontier of end-to-end document understanding.

\subsection{Performance on Multi-Page Benchmarks}
We report the performance of existing multi-page frameworks on two multi-page VRDU benchmarks in Table~\ref{tab:multipage_performance}. General-domain models can achieve reasonable performance; however, frameworks equipped with mechanisms explicitly designed for visually rich documents (VRDs) consistently yield substantial improvements. Currently, most high-performing multi-page methods rely on OCR-dependent pipelines and achieve strong results by leveraging external OCR tools. While such designs reduce the burden of directly understanding and compressing visual representations, they also inherit the limitations of OCR-based approaches, including error accumulation as observed in single-page scenarios. For multi-page tasks, this challenge is further amplified, highlighting the need for more effective strategies to manage the large number of visual tokens and to improve text understanding in multi-page, text-dense document inputs.

\newpage
\begin{table*}[!h]
\definecolor{mya}{HTML}{0039a6}
\definecolor{myb}{HTML}{fccc0a}

\newcommand{\trainable}{\textcolor{green!60!black}{\ding{51}}\xspace} % ✓
\newcommand{\frozen}{\textcolor{red!70!black}{\ding{55}}\xspace}    % ✗
\centering
\small
%\begin{adjustbox}{max width=\textwidth}
\begin{tabular}{l|ccc|ccc|ccc}
\toprule
  & \multicolumn{3}{c|}{\textbf{Vision Encoder}} 
  & \multicolumn{3}{c|}{\textbf{LLM Backbone}} 
  & \multicolumn{3}{c}{\textbf{Adaptors}} \\ 
\cmidrule(lr){2-4} \cmidrule(lr){5-7} \cmidrule(lr){8-10}  
\textbf{Model Name}  & PT & IT & SFT 
  & PT & IT & SFT 
  & PT & IT & SFT \\
\midrule
\rowcolor[gray]{.9}\multicolumn{10}{l}{\textbf{OCR-Dependent}}\\
ICL-D3IE        \shortcite{icld3ie}       & --       & --       & --       & --       & --       & --       & --       & --       & --       \\
DocLLM         \shortcite{docllm}        & \trainable & \trainable & --       & --       & --       & --       & \trainable & \trainable & --       \\
LAPDoc         \shortcite{lapdoc}        & --       & --       & --       & --       & --       & --       & --       & --       & --       \\
LMDX           \shortcite{lmdx}          & --       & --       & --       & --       & --       & --       & --       & --       & --       \\
ProcTag        \shortcite{proctag}       & --       & --       & \trainable & --       & --       & \trainable & --       & --       & \trainable\\
DocKD          \shortcite{dockd}         & --       & --       & \trainable & --       & --       & \trainable & --       & --       & --       \\
DoCo           \shortcite{li2024enhancing} & \frozen  & --       & \frozen    & \trainable & --       & \frozen    & \trainable & --   &    \trainable \\
InstructDoc     \shortcite{instructdoc}   & --       & \frozen    & \frozen    & --       & \frozen    & \frozen    & --       & \trainable & \trainable \\
LayoutLLM      \shortcite{luo2024layoutllm} & --    & \frozen    & \trainable & --       & \frozen    & \frozen    & --       & \trainable & \trainable \\
LLaVA-Read    \shortcite{llavaread}     & \frozen    & \trainable & --       & \frozen    & \frozen    & --       & \trainable & \trainable & --       \\
LayTextLLM     \shortcite{lu2024bounding} & \frozen   & --       & \trainable & --       & --       & --       & \trainable & --       & \trainable \\
LayTokenLLM    \shortcite{laytokenllm} & \frozen    & --       & \frozen    & --       & --       & --       & \trainable & --       & \trainable \\
GPE            \shortcite{gpe}           & --       & --       & \trainable & --       & --       & --       & --       & --       & --       \\
MDocAgent      \shortcite{mdocagent}     & --       & --       & --       & --       & --       & --       & --       & --       & --       \\
PDF-WuKong     \shortcite{pdfwukong}     & --       & --       & \trainable & --       & --       & \trainable & --       & --       & --       \\
DocLayLLM      \shortcite{doclayllm}     & \frozen    & \frozen    & --       & \trainable & \trainable & --       & \trainable & \trainable & --       \\
DocAssistant    \shortcite{zhang2025docassistant} & - & \frozen & - & - & \frozen & - & - & \trainable & - \\
AlignVLM \shortcite{masryalignvlm} & \trainable & \trainable & \frozen & \trainable & \trainable & \trainable & \trainable & \trainable & \trainable \\
DocThinker \shortcite{yu2025docthinker} & - & - & \trainable & - & - & \trainable & - & - & \trainable\\
\midrule
\rowcolor[gray]{.9}\multicolumn{10}{l}{\textbf{OCR-Free}}\\
KOSMOS-2.5     \shortcite{lv2023kosmos}  & --       & \trainable & \trainable & --       & \trainable & \frozen    & --       & \trainable & \trainable \\
mPLUG-DocOwl   \shortcite{mplugdocowl}   & --       & \frozen    & --       & --       & \frozen    & --       & --       & \trainable & -- \\
UReader        \shortcite{ureader}       & --       & \frozen    & --       & --       & \frozen    & --       & --       & \trainable & --       \\
TGDoc          \shortcite{wang2023towards} & --     & \frozen    & \trainable & --       & \frozen    & \frozen    & --       & \trainable & \trainable \\
UniDoc         \shortcite{unidoc}        & --       & \frozen    & \trainable & --       & \frozen    & \frozen    & --       & \trainable & \trainable \\
DocPedia       \shortcite{feng2024docpedia} & \frozen & --       & \trainable & \trainable & --       & \trainable & \trainable & --       & \trainable \\
HRVDA          \shortcite{hrvda}         & \frozen    & \frozen    & --       & \trainable & \frozen    & --       & \trainable & \trainable & --       \\
Vary           \shortcite{vary}          & \trainable & --       & \trainable & {\trainable} & -- & \frozen & \trainable & -- & \trainable \\
mPLUG-DocOwl 1.5 \shortcite{mplugdocowl15} & --      & \frozen    & \trainable & --       & \trainable & \frozen    & --       & \trainable & \trainable \\
HVFA           \shortcite{park2024hierarchical} & -- & \frozen    & --       & --       & \frozen    & --       & --       & {\trainable} & -- \\
mPLUG-DocOwl2  \shortcite{mplugdocowl2}  & --       & \frozen    & \trainable & --       & \trainable & \frozen    & --       & \trainable & \trainable \\
Texthawk       \shortcite{texthawk}      & --       & \frozen    & \trainable & --       & \frozen    & \frozen    & --       & \trainable & \trainable \\
Texthawk2      \shortcite{texthawk2}     & --       & \frozen    & \trainable & --       & \frozen    & \trainable & --       & \trainable & \trainable \\
TextMonkey     \shortcite{textmonkey}    & --       & \trainable & --       & --       & \trainable & --       & \trainable & \trainable & --       \\
Llavar         \shortcite{llavar}        & --       & \frozen    & \trainable & --       & \frozen    & \frozen    & --       & \trainable & \trainable \\
TokenCorrCompressor \shortcite{zhangtoken} & --    & --       & \frozen    & --       & --       & \frozen    & --       & --       & \trainable \\
DocKylin       \shortcite{dockylin}      & --       & \frozen    & \trainable & --       & \trainable & \trainable & --       & \trainable & \trainable \\
Marten        \shortcite{marten}        & --       & \frozen    & \trainable & --       & \trainable & \trainable & --       & \trainable & \trainable \\
PP-DocBee      \shortcite{ppdocbee}      & --       & --       & \trainable & --       & --       & \frozen    & --       & --       & --       \\
TokenFD \shortcite{guan2025token} & \trainable & \frozen & \trainable & \trainable & \frozen & \trainable & \trainable & \trainable & \trainable \\
\bottomrule
\end{tabular}
%\end{adjustbox}
\caption{Comparison of MLLM-based VRDU frameworks. PT~-~Pretraining, IT~-~Instruction-tuning, SFT~-~Supervised Fine-tuning. 
}
%-- represents no relevant training stage. }
\label{tab:mlmm-vrdu-training}
\end{table*}

\newpage
\begin{table*}[t]
\centering
\scriptsize

\begin{tabular}{p{2.2cm} p{2.5cm} p{2.2cm} p{1.8cm}}
\toprule
\textbf{Model} & \textbf{LLM Backbone} & \textbf{Vision} & \textbf{Adaptor} \\
\midrule

\multicolumn{4}{c}{\textbf{OCR-Dependent}} \\
\midrule
ICL-D3IE & GPT-3, ChatGPT & -- & -- \\
DocLLM & Falcon-1B / LLaMA2-7B & -- & Spatial Attention \\
LAPDoc & ChatGPT, Solar & -- & -- \\
LMDX & PaLM2-S, Gemini Pro & -- & -- \\
ProcTag & Qwen-7B / Qwen-VL & Qwen2VL & Projector \\
DocKD & DocFormerV2 & DocFormerV2 & -- \\
DoCo & Qwen-VL / mPLUG-Owl & ViT-bigG & VL Adapter \\
InstructDr & Flan-T5 & CLIP & DocFormer \\
LayoutLLM & Vicuna / LLaMA2 & LayoutLMv3 & MLP \\
LLaVA-Read & Vicuna-13B & CLIP-ViT-L & MLP \\
LayTextLLM & LLaMA2-7B & -- & Layout LoRA \\
LayTokenLLM & Qwen / LLaMA3 & -- & Layout Tokenizer \\
GPE & LLaMA2 / Qwen & -- & -- \\
MDocAgent & LLaMA3 / Qwen2-VL & ColPali & -- \\
PDF-WuKong & IXC2-VL & IXC2-VL & -- \\
DocLayLLM & LLaMA2 / LLaMA3 & LayoutLMv3 & Projector + LoRA \\
DocAssistant & InternVL2 & InternVL2 & MoM Adapter \\
AlignVLM & Llama3.1 & SigLIP & ALIGN \\
DocThinker & Qwen2.5-VL & Qwen2.5-VL & -- \\

\midrule
\multicolumn{4}{c}{\textbf{OCR-Free}} \\
\midrule
KOSMOS-2.5 & Transformer & Pix2Struct & Resampler \\
DocOwl & mPLUG-Owl & ViT & Abstractor \\
UReader & mPLUG-Owl & CLIP-ViT & Abstractor \\
TGDoc & Vicuna-7B & CLIP & MLP \\
UniDoc & Vicuna & CLIP & MLP \\
DocPedia & Vicuna-7B & Swin & MLP \\
HRVDA & LLaMA2 & Swin & Detector + LoRA \\
Vary & OPT + Qwen & CLIP + SAM & MLP \\
DocOwl1.5 & mPLUG-Owl2 & ViT & Reducer \\
HVFA & BLIP2 / mPLUG & ViT & HVFA + LoRA \\
DocOwl2 & mPLUG-Owl2 & ViT & Reducer \\
Texthawk & InternLM & SigLIP & Resampler \\
Texthawk2 & Qwen2 & SigLIP & Multi-module \\
TextMonkey & Qwen-VL & ViT-BigG & Resampler \\
LLaVAR & Vicuna-13B & CLIP & MLP \\
TokenCorr & LLaMA2 & CLIP & Compressor \\
DocKylin & Qwen-7B & Swin & MLP + APS \\
Marten & InternLM2 & InternViT & Mask Module \\
PP-DocBee & Qwen2-VL & ViT & -- \\
TokenFD & Embedding & ViT & Abstractor \\

\bottomrule
\end{tabular}
\caption{MLLM-based VRDU frameworks.}
\label{tab:mlmm-vrdu-backbones}
\end{table*}
\newpage

\begin{table*}[ht]
\centering
\scriptsize
\setlength{\tabcolsep}{3pt}
\begin{adjustbox}{max width=\textwidth}
\begin{tabular}{l l l l l l l l l l}
\toprule
\textbf{Tool Name} & \textbf{Provider} & \textbf{Tool Type} & \textbf{Deployment} & \textbf{Pricing} & \textbf{Input Modalities} & \textbf{Languages} & \textbf{Openness} & \textbf{GPU} & \textbf{Doc Parsing} \\
\midrule
pdfminer.six & Y. Shinyama et al. & OCR Engine & Local & Free & PDF & Multi & Open-source & No & No \\
Mistral OCR & Mistral AI & Document API & Cloud & Paid (Usage-based) & Image, PDF & Multi & Closed & Supported & Yes \\
LightOnOCR & LightOnAI & Vision LLM & Cloud & Paid (Usage-based) & Image, PDF & Multi & Closed & Supported & No \\
Google Cloud Vision & Google & Document API & Cloud & Paid (Usage-based) & Image, PDF & Multi & Closed & Supported & Yes \\
Kraken & Inria et al. & OCR Engine & Local & Free & Image, PDF & Multi & Open-source & Supported & No \\
Qwen3-VL & Aliyun & Vision LLM & Hybrid & Free* & Image, PDF & Pretrained/Dependent & Closed & Supported & No \\
olmOCR & AI2 & OCR Engine & Hybrid & Free & Image, PDF & Multi & Open-source & Supported & Yes \\
AttentionOCR & Guo \& Deng & OCR Engine & Local & Free & Image & Multi & Open-source & Supported & No \\
Calamari & Univ. Würzburg & OCR Engine & Local & Free & Image & Multi & Open-source & Supported & No \\
EasyOCR & JaidedAI & OCR Engine & Local & Free & Image & Multi & Open-source & Supported & No \\
OpenAI Vision & OpenAI & Vision LLM & Cloud & Paid (Usage-based) & Image, PDF & Multi & Closed & Supported & Yes \\
Tesseract & S. Weil & OCR Engine & Local & Free & Image & Multi & Open-source & No & No \\
Adobe PDF Extract & Adobe & Document API & Cloud & Paid (Usage-based) & PDF & Multi & Closed & Supported & Yes \\
PaddleOCR & PaddlePaddle & OCR Engine & Cloud & Free & Image, PDF & Multi & Open-source & Supported & Yes \\
docTR & Mindee & OCR Engine & Local & Free & Image, PDF & Pretrained/Dependent & Open-source & Supported & No \\
DeepSeek-OCR & DeepSeek AI & Vision LLM & Hybrid & Paid (Usage-based) & Image, PDF & Multi & Open-source & Supported & No \\
HunyuanOCR & Tencent & Vision LLM & Local & Free & Image, PDF & Multi & Open-source & Supported & No \\
Ocular & Berkeley NLP & OCR Engine & Local & Free & Image, PDF & Multi & Open-source & Supported & No \\
MinerU & OpenDataLab & Document API & Local & Free & PDF & Multi & Open-source & Supported & Yes \\
SuryaOCR & Datalab & OCR Engine & Local & Free & Image, PDF, Word, PPT & Multi & Open-source & Supported & No \\
Seed-VL & ByteDance Seed & Vision LLM & Cloud & Paid (Usage-based) & Image, PDF & Multi & Open-source & Supported & Yes \\
\bottomrule
\end{tabular}
\end{adjustbox}
\caption{Comparison of OCR engines, document parsing APIs, and vision-language models for document understanding.}
\label{tab:ocr_llm_comparison}
\end{table*}
\newpage
\begin{table*}[!h]
\centering
\footnotesize
%\rowcolors{2}{}{gray!10}
%\begin{adjustbox}{max width=\linewidth}
\begin{tabularx}{\textwidth}{l l X r c}
\toprule
\textbf{Study} & \textbf{Dataset} & \textbf{Source} & \textbf{Size} & \makecell[bc]{\textbf{Public}\\ \textbf{Available}} \\
\midrule
Vary
 & Document Data Engine & ArXiv, CC-MAIN, E-books & 2M & \xmark \\
 & Chart Data Engine & matplotlib, pyecharts, NLP corpora & 1.5M & \xmark \\
 & Detection Data Engine & Objects365, OpenImages & $\sim$3M & \cmark\\
%\midrule
LLaVAR & LAION & LAION images filtered for text-rich content, OCR applied & 0.4M & \cmark \\
%\midrule
DoCo & DoCo-Processed & CC3M (LLaVA) + LAION, processed with PaddleOCR & 1.0M & \xmark \\
%\midrule
Texthawk2 & 100M pretraining & Diverse, mainly public datasets & 100M & \xmark \\
%\midrule
Docpedia
 & PDF Images & arXiv (public scientific preprints) & 325K & \cmark \\
 & PPT Images & Common Crawl (web-crawled PPTs) & 600K & Partly \\
\bottomrule
\end{tabularx}
%\end{adjustbox}
\caption{Summary of pretraining datasets created and used in recent MLLM-based VRDU frameworks.}
\label{tab:pretraining_datasets}
\end{table*}

\newpage
\begin{table*}[!h]
\centering
%\begin{adjustbox}{max width=\linewidth}
\small
%\rowcolors{2}{}{gray!10}
\begin{tabularx}{\textwidth}{l>{\raggedright\arraybackslash}p{8em}>{\raggedright\arraybackslash}Xrl}
\toprule
\textbf{Framework} & \textbf{Category} & \textbf{Source / Description} & \textbf{Size (K)} & \textbf{Open Source} \\
\midrule
{Leopard}
    & Multi-image (text-rich) 
    & 69K public multi-page docs/slides; Adapted single-page to multi-image (DocVQA, ArxivQA); Raw slides + GPT-4o QAs; Multi-chart/table (open, synth.); Webpage snapshots (Mind2Web, OmniACT, WebScreenshots, etc.)
    & 739 & Partially \\
    \cmidrule{2-5}
    & Single-image 
    & Text-rich single images from public datasets; Natural images (e.g., ShareGPT4V, etc.)
    & 186 & Partially \\
\midrule
{LLaVAR}
    & Noisy Instruction-Following 
    & Text-rich images from LAION, selected via classifier + CLIP clustering, instructions via OCR-based prompts
    & 422,000 & Yes \\
    \cmidrule{2-5}
    & High-Quality Instruction-Following 
    & Subset of LAION text-rich images (4 clusters), multi-turn QAs generated by prompting text-only GPT-4 with OCR+caption info
    & 16,000 & Yes \\
\bottomrule
\end{tabularx}
%\end{adjustbox}
\caption{Summary of instruction-tuning datasets for Leopard and LLaVAR.}
%\textsuperscript{*}Partially: only part of the data is open.}
\label{tab:instruction_tuning_dataset}
\end{table*}

\newpage
\begin{table*}[!h]
\centering
\setlength{\tabcolsep}{3pt}
\scriptsize
%\begin{adjustbox}{max width=\linewidth}
\begin{tabular}{llllrrrclll}
\toprule
\textbf{Dataset} & \textbf{Venue} & \textbf{Year} & \textbf{Domain} & \textbf{Docs} & \textbf{Images} & \textbf{Keys / Qs} & \textbf{Multi page} & \textbf{Language} & \textbf{Metrics} & \textbf{Format}\\
\midrule
\rowcolor[gray]{.9}\multicolumn{11}{l}{\textbf{Key Information Extraction}}\\
FUNSD & ICDAR-w & 2019 & Multi-source & -- & 199 & 4 & \xmark & English & F1 & P, H\\
SROIE & ICDAR-c & 2019 & Scanned Receipts & -- & 973 & 4 & \xmark & English & F1* & P\\
CORD& NeurIPS-w & 2019 & Scanned Receipts & -- & 1,000 & 54 & \xmark & English & F1 & P\\
Payment-Invoice & ACL & 2020 & Invoice Form & -- & 14,832 & 7 & \xmark & English & F1 & D\\
Payment-Receipts & ACL & 2020 & Scanned Receipts & -- & 478 & 2 & \xmark & English & F1 & P\\
Kleister-NDA & ICDAR & 2021 & Private Agreements & 540 & 3,229 & 4 & \cmark & English & F1 & D\\
Kleister-Charity & ICDAR & 2021 & AFR & 2,778 & 61,643 & 8 & \cmark & English & F1 & D, P\\
EPHOIE & AAAI & 2021 & Exam Paper & -- & 1,494 & 10 & \xmark & Chinese & F1 & P, H\\
XFUND & ACL & 2022 & Synthetic Forms & -- & 1,393 & 4 & \xmark & Multilingual & F1 & D, P, H\\
Form-NLU & SIGIR & 2023 & Financial Form & -- & 857 & 12 & \xmark & English & F1 & D, P, H\\
VRDU-Regist. Form & KDD & 2023 & Registration Form & -- & 1,915 & 6 & \xmark & English & F1 & D\\
VRDU-Ad-buy Form & KDD & 2023 & Political Invoice Form & -- & 641 & 9+1(5) & \xmark & English & F1 & D, P\\
DocILE & ICDAR & 2023 & Invoice Form & 6,680 & 106,680 & 55 & \cmark & English & AP, CLEval & D, P\\
KVP10k & ICDAR & 2024 & Cross-domain & -- & 10,707 & 118,868 & \xmark & English &  F1, IOU & D, H \\
CC-OCR-KIE & ICCV & 2025 & Cross-domain & -- & 2,008 &  34(-) & \xmark & Multilingual & F1 & D, P, H \\
\midrule
\rowcolor[gray]{.9}\multicolumn{11}{l}{\textbf{Visual Question Answering}}\\
DocVQA & WACV & 2021 & Industrial Reports & -- & 12,767 & 50,000 & \xmark & English & ANLS & D, P, H\\
VisualMRC & AAAI & 2021 & Website & -- & 10,197 & 30,562 & \xmark & English & BLEU, etc & D\\
TAT-DQA & MM & 2022 & Financial Reports & 2,758 & 3,067 & 16,558 & \cmark & English & EM, F1 & D\\
RDVQA & MM & 2022 & Data Analysis Report & 8,362 & 8,514 & 41,378 & \xmark & English & ANLS, ACC & D\\
CS-DVQA & MM & 2022 & Industry Documents & -- & 600 & 1,000 & \xmark & English & ANLS & D, P, H\\
InfographicVQA & WACV & 2022 & Infographics & -- & 5,400 & 3,000 & \xmark & English & ANLS, F1 & D \\
PDFVQA-Task A & ECML-PKDD & 2023 & Academic Paper & -- & 12,337 & 81,085 & \xmark & English & F1 & D\\
PDFVQA-Task B & ECML-PKDD & 2023 & Academic Paper & -- & 12,337 & 53,872 & \xmark & English & F1 & D\\
PDFVQA-Task C & ECML-PKDD & 2023 & Academic Paper & 1,147 & 12,337 & 5,653 & \cmark & English & EM & D\\
MPDocVQA & PR & 2023 & Industrial Reports & 6,000 & 48,000 & 46,000 & \cmark & English & ANLS & D, P, H\\
DUDE & ICCV & 2023 & Cross-domain & 5,019 & 28,709 & 41,541 & \cmark & English & ANLS & D\\
SlideVQA & AAAI & 2023 & Slide, decks & -- & 5,200 & 14,500 & \cmark & English & EM, F1 & D \\
MMLONGBENCH-DOC & NIPS & 2024 & Cross-domain & 135 & 6,413 & 1,082 & \cmark & English & ACC, F1 & D \\
MMVQA & IJCAI & 2024 & Academic Paper & 3,146 & 30,239 & 262,928 & \cmark & English & EM, PM, MR & D\\
JDocQA & LREC-COLING & 2024 & Cross-Domain & 5,504 & 268,000 & 11,600 & \cmark & Japanese & F1 & D \\
BoundingDocs & IJDAR & 2025 & Cross-domain, Mixed  & 48,151 & 237,437 & 249,016 & \xmark & Multilingual & ANLS & D, P, H\\
LongDocURL & ACL & 2025 & Cross-domain & 396 & 33,000 & 2,325 & \cmark & English & F1 & D \\
MMDocIR & EMNLP & 2025 & Cross-domain & 6,878 & 224,223 & 73,843 & \cmark & Multilingual & F1 & D \\
MTVQA & EMNLP & 2025 & Cross-domain & - & 8,794 & 28,607 & \xmark & Multilingual & ANLS & D, P, H \\
SEEPHYS & NIPS & 2025 & Physics & - & 2,245 & 2,000 & \xmark & English & Accuracy & D \\
\bottomrule
\end{tabular}
%\end{adjustbox}
\caption{Benchmark datasets for Key Information Extraction and Visual Question Answering in visually rich documents. P - Scanned \textbf{P}rinted, H - Scanned \textbf{H}andwritten, D - \textbf{D}igtial Born}
\label{tab:vrd_benchmark_dataset}
\end{table*}

\newpage
\begin{table*}[ht]
\centering
\small
\begin{tabular}{lrrrrrr}
  \toprule
%\multirow{2}{*}{\textbf{Model Name}} & \multicolumn{6}{c}{\textbf{Dataset}} \\ \cmidrule{2-7}
\textbf{Model Name} & \textbf{FUNSD} & \textbf{CORD} & \textbf{SROIE} & \textbf{DocVQA} & \textbf{ChartVQA} & \textbf{InfoVQA} \\   
\midrule

\rowcolor[gray]{.9}\textbf{General Domain LLM}&&&&&&\\
Qwen1.5-7B-Chat & 52.5 & 29.7 & -- & 64.3 & -- & -- \\
Llama3-8B-Instruct & 57.5 & 40.0 & -- & 74.2 & -- & -- \\
\midrule

\rowcolor[gray]{.9}\textbf{General Domain MLLM}&&&&&&\\

QwenVL-7B & 47.1 & 30.0 & -- & 65.1 & -- & -- \\
InterVL2-8B & 75.8 & 79.9 & -- & 91.7 & -- & -- \\
Claude-3.5 Sonnet & -- & -- & -- & 88.5 & 51.8 & 59.1 \\
GeminiPro-1.5 & -- & -- & -- & 91.2 & 34.7 & 73.9 \\
GPT4o 20240806 & -- & -- & -- & 92.8 & 85.7 & 66.4 \\

\midrule
\rowcolor[gray]{.9}\textbf{OCR-Dependent}&&&&&&\\
DocLLM \shortcite{docllm}       & 51.8  & 67.4  & 91.9  & 69.5  & --     & --     \\ 
LAPDoc \shortcite{lapdoc}       & --     & --     & --     & 79.8  & --     & 54.9  \\ 
DoCo \shortcite{li2024enhancing}& --     & --     & --     & 64.8  & 68.9  & 34.9  \\ 
InstructDr \shortcite{instructdoc}   & 38.1  & 62.7  & --     & 22.3  & --     & 37.6  \\ 
LayoutLLM \shortcite{luo2024layoutllm}  & 78.7  & 62.2  & 71.0    & 74.3  & --     & --     \\ 
LLaVA-Read \shortcite{llavaread}     & 36.9  & --     & 58.3  & 71.0    & 74.6  & 36.4  \\ 
LayTextLLM \shortcite{lu2024bounding}  & 64.0    & 96.5  & 95.8  & 77.2  & --     & --     \\ 
LayTokenLLM\shortcite{laytokenllm}   & 71.0 & 75.4 & --     & 85.1 & --     & --     \\ 
GPE \shortcite{gpe}                & 82.6  & 86.9  & 97.8  & 78.1  & --     & --     \\ 
PDF-WuKong \shortcite{pdfwukong}    & 85.1  & --     & --     & 76.9  & 80.0    & 61.3  \\ 
DocLayLLM \shortcite{doclayllm}     & 80.7  & 79.4  & 84.4  & 72.8  & --     & --     \\
AlignVLM \shortcite{masryalignvlm} & -- & -- & -- & 81.2 & 75.0 & 53.8\\
DocAssistant    \shortcite{zhang2025docassistant}  & -- & -- & -- & 89.8 & 81.4 & 66.7\\
DocThinker \shortcite{yu2025docthinker}  & 	- & -- & 81.4 & 80.2 & -- & 69.7\\
\midrule
\rowcolor[gray]{.9}\textbf{OCR-Free}&&&&&&\\
KOSMOS-2.5~\shortcite{lv2023kosmos}      & --     & --     & --     & 81.1  & 62.3  & 41.3  \\ 
mPLUG-DocOwl~\shortcite{mplugdocowl2}    & --     & --     & --     & 62.2  & 57.4  & 38.2  \\ 
UReader~\shortcite{ureader}              & --     & --     & --     & 65.4  & 59.3  & 42.2  \\ 
TGDoc~\shortcite{wang2023towards}        & 1.7   & --     & 3.0     & 9.0    & 11.7 & 12.8 \\ 
UniDoc~\shortcite{unidoc}                & 1.2  & --     & 1.4   & 6.5  & 10.5 & 13.8 \\ 
DocPedia~\shortcite{feng2024docpedia}    & 40.1  & --     & 57.7  & 49.3  & 47.8  & 15.5  \\ 
HRVDA~\shortcite{hrvda}                  & --     & 89.3  & 89.3  & 91.0  & 72.1  & 43.5  \\ 
Vary-base~\shortcite{vary}               & --     & --     & --     & 76.3  & 66.1  & --     \\ 
mPLUG-DocOwl 1.5~\shortcite{mplugdocowl15} & --   & --     & --     & 81.6  & 70.5  & 50.4  \\ 
HVFA~\shortcite{park2024hierarchical}    & --     & --     & --     & 72.7  & 63.3  & 45.9  \\ 
mPLUG-DocOwl2~\shortcite{mplugdocowl2}   & --     & --     & --     & 80.7  & 70.0   & 46.4  \\ 
Texthawk~\shortcite{texthawk}            & --     & --     & --     & 76.4  & 66.6  & 50.6  \\ 
Texthawk2~\shortcite{texthawk2}          & --     & --     & --     & 89.6  & 81.4  & 67.8  \\ 
TextMonkey~\shortcite{textmonkey}        & 65.5 & 67.5 & 47.0    & 73.0    & 66.9  & 28.6  \\ 
Llavar-7B~\shortcite{llavar}             & 1.7  & 13.6 & 2.4  & 11.6  & --     & --     \\ 
TokenCorrCompressor~\shortcite{zhangtoken} & --    & --     & --     & 78.3  & 68.9  & 50.2  \\ 
DocKylin~\shortcite{dockylin}            & 25.5  & --     & 49.5  & 77.3  & 66.8  & 46.6  \\ 
Marten~\shortcite{marten}                & 44.4  & --     & 80.4  & 92.0    & 81.7  & 75.2  \\ 
PP-DocBee~\shortcite{ppdocbee}           & --     & --     & --     & 90.6  & 74.6  & 66.2  \\   
TokenFD \shortcite{guan2025token} & 42.2 & -- & 81.9 & 94.2 & 86.6 & 76.5\\
\bottomrule
\end{tabular}
\caption{Performance comparison between OCR-dependent and OCR-free document understanding frameworks across benchmark datasets.}
\label{tab:dataset_comparison}
\end{table*}

\newpage
\begin{table*}[!h]
%\definecolor{lightgray}{gray}{0.95}
\definecolor{best}{RGB}{220, 20, 60} % crimson for best values
\centering
%\renewcommand{\arraystretch}{1.25}
%\setlength{\tabcolsep}{6pt}
%\begin{adjustbox}{max width=0.6\textwidth}
\small
\begin{tabular}{lllrrr}
\toprule
\textbf{Model} & \textbf{Type} & \textbf{Venue} & \textbf{Year} & \textbf{MPDocVQA} & \textbf{DUDE} \\
\midrule
Longformer & General VLPM & Preprint & 2020 & 55.1 & 20.3 \\
BigBird & General VLPM & NeurIPS & 2020 & 58.5 & 26.3 \\
GPT-4v & General MLLM & -- & 2023 & -- & 53.9 \\
Idefics3-8B & General MLLM & Preprint & 2024 & 67.2 & 38.7 \\
LLaVA-next-interleave-7B & General MLLM & Preprint & 2024 & 44.9 & 28.0 \\

Hi-VT5 &  OCR-dependent VLPM & PR & 2023 & 61.8 & 35.7 \\
GRAM & OCR-dependent VLPM & CVPR & 2024 & \textbf{\textcolor{best}{83.0}} & 53.4 \\
InstructDoc & OCR-Dependent VLPM & AAAI & 2024 & -- & 46.8 \\
mPLUG-DocOwl2 & OCR-free VLPM & Preprint & 2024 & 69.4 & 46.7 \\
PDF-WuKong & OCR-Dependent VLPM & Preprint & 2024 & 76.9 & 56.1 \\
LayTokenLLM & OCR-Dependent VLPM & CVPR & 2025 & 74.3 & 52.0 \\
DocThinker \shortcite{yu2025docthinker} & OCR-Dependent VLPM & ICCV & 2025 & -- & \textbf{\textcolor{best}{56.8}}\\
\bottomrule
\end{tabular}
%\end{adjustbox}
\caption{Performance comparison of state-of-the-art models on MPDocVQA and DUDE benchmarks. Best scores are highlighted in \textcolor{best}{red}.}
\label{tab:multipage_performance}
\end{table*}

\end{document}